\documentclass[11pt,a4paper]{article}
\usepackage[hyperref]{acl2021}
\usepackage{times}
\usepackage{latexsym}

\usepackage{epsfig}
\usepackage{graphicx}
\usepackage{subfigure}
\usepackage{overpic}
\usepackage{amsfonts}
\usepackage{booktabs}
\usepackage{bm}
\usepackage{amsmath}
\usepackage{bbm}
\usepackage{url}

\newcommand{\figref}[1]{Figure~\ref{#1}}
\newcommand{\tabref}[1]{Table~\ref{#1}}
\newcommand{\equref}[1]{Eq.~(\ref{#1})}

\usepackage{booktabs}
\usepackage{threeparttable}
\usepackage{multirow}

\usepackage{microtype}

\aclfinalcopy 
\def\aclpaperid{384} 

\title{Multimodal Incremental Transformer with Visual Grounding \\ for Visual Dialogue Generation}

\author{{Feilong Chen}, {Fandong Meng}, {Xiuyi Chen}, {Peng Li},  {Jie Zhou} \\
  Pattern Recognition Center, WeChat AI, Tencent Inc, Beijing, China \\
  {\tt
  \{\href{mailto:ivess.chan@gmail.com}{ivess.chan},\href{mailto:hugheren.chan@gmail.com
}{hugheren.chan}\}@gmail.com}\\
  {\tt \{\href{mailto:fandongmeng@tencent.com}{fandongmeng},\href{mailto:patrickpli@tencent.com}{patrickpli},\href{mailto:withtomzhou@tencent.com}{withtomzhou}\}@tencent.com}\\
  }

\date{}

\begin{document}
\maketitle
\begin{abstract}
Visual dialogue is a challenging task since it needs to answer a series of coherent questions on the basis of understanding the visual environment. Previous studies focus on the implicit exploration of multimodal co-reference by implicitly attending to spatial image features or object-level image features but neglect the importance of locating the objects explicitly in the visual content, which is associated with entities in the textual content. Therefore, in this paper we propose a {\bf M}ultimodal {\bf I}ncremental {\bf T}ransformer with {\bf V}isual {\bf G}rounding, named MITVG, which consists of two key parts: visual grounding and multimodal incremental transformer. Visual grounding aims to explicitly locate related objects in the image guided by textual entities, which helps the model exclude the visual content that does not need attention. On the basis of visual grounding, the multimodal incremental transformer encodes the multi-turn dialogue history combined with visual scene step by step according to the order of the dialogue and then generates a contextually and visually coherent response. Experimental results on the VisDial v0.9 and v1.0 datasets demonstrate the superiority of the proposed model, which achieves comparable performance.
\end{abstract}

\section{Introduction}

\begin{figure}[t!]
\centering
\scalebox{0.98}{
  \begin{overpic}[width=\columnwidth]{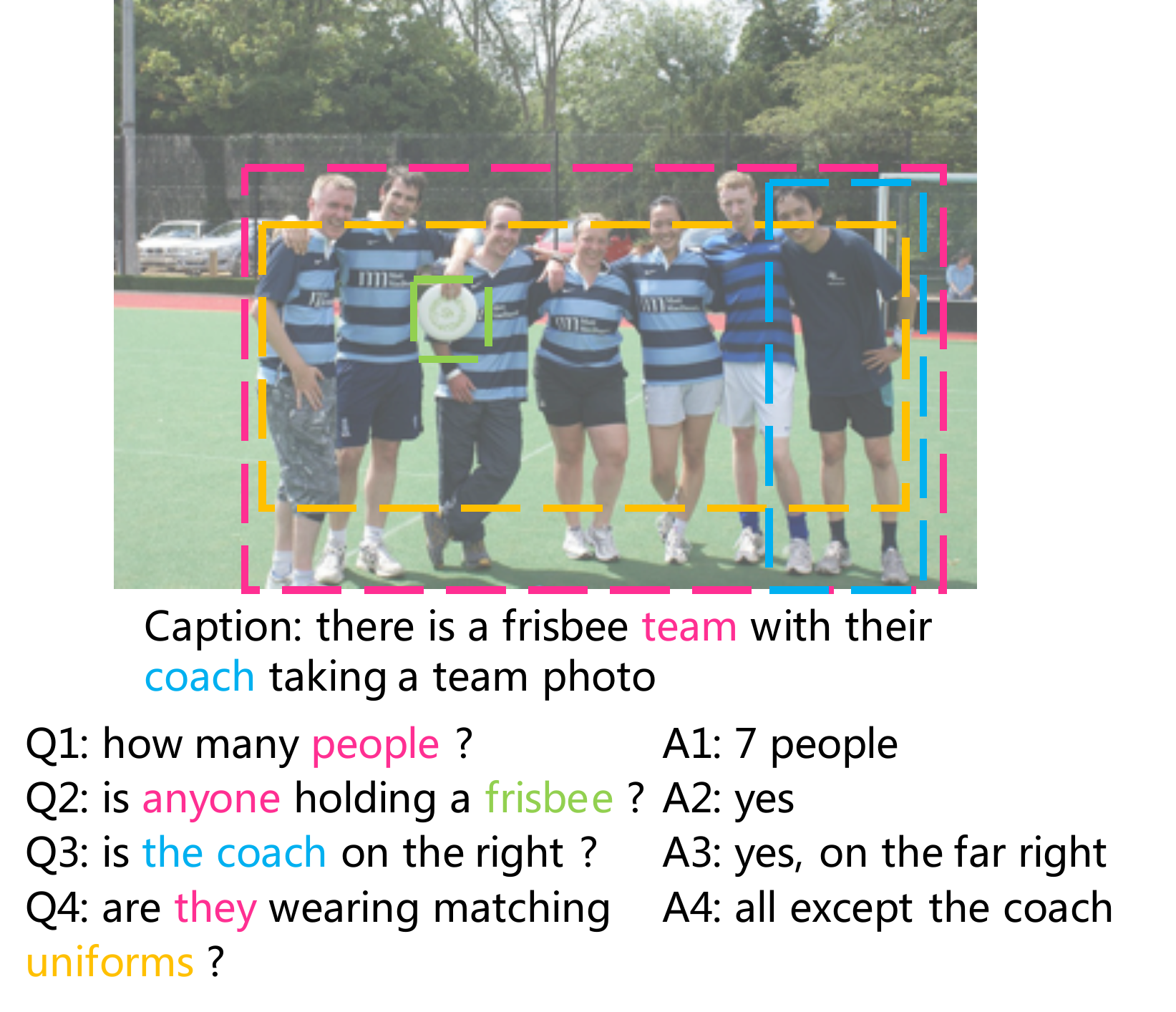}
  \end{overpic}
  }
  \caption{An example of visual dialogue. The color in text background corresponds to the same color box in the image, which indicates the same entity. Our model firstly associates textual entities with objects explicitly and then gives contextually and visually coherent answers to contextual questions.
  }\label{fig:example}
\end{figure}

Recently, there is increasing interest in vision-language tasks, such as image caption~\cite{xu2015show,Anderson2016SPICE,anderson2018bottom,cornia2020m2} and visual question answering~\cite{ren2015exploring,gao2015you,lu2016hierarchical,anderson2018bottom}. In the real world, our conversations~\cite{chen2020bridging,chen2019working} usually have multiple turns. 
As an extension of conventional single-turn visual question answering, \citet{das2017visual} introduce a multi-turn visual question answering task named visual dialogue, which aims to explore the ability of an AI agent to hold a meaningful multi-turn dialogue with humans in natural language about visual content.

Visual dialogue~\cite{agarwal2020history,wang2020vd,Qi2020TwoCP,murahari2019large} requires agents to give a response on the basis of understanding both visual and textual content. One of the key challenges in visual dialogue is how to solve multimodal co-reference~\cite{das2017visual,Kottur2018VisualCR}. 
Therefore, some fusion-based models~\cite{das2017visual} are proposed to fuse spatial image features and textual features in order to obtain a joint representation. Then attention-based models~\cite{lu2017best,wu2018you,Kottur2018VisualCR} are proposed to dynamically attend to spatial image features in order to find related visual content. Furthermore, models based on object-level image features~\cite{niu2019recursive,gan2019multi,chen2020dmrm,jiang2020kbgn,nguyenefficient,jiang2020dam} are proposed to effectively leverage the visual content for multimodal co-reference. However, as implicit exploration of multimodal co-reference, these methods implicitly attend to spatial or object-level image features, which is trained with the whole model and is inevitably distracted by unnecessary visual content. Intuitively, specific mapping of objects and textual entities can reduce the noise of attention.
As shown in \figref{fig:example}, the related objects can help the agent to understand the entities (e.g., Q1: ``{\em people}'', Q2: ``{\em frisbee}'', Q3: ``{\em coach}'') for the generation of correct answers. Then when it answers the question Q4 ``{\em are they wearing matching uniforms ?}'', the agent has already comprehended ``{\em people}'' and ``{\em coach}'' from the previous conversation. On this basis, it can learn the entity ``{\em uniforms}'' with the corresponding object in the image, and generate the answer ``{\em all except the coach}''.
To this end, we need to 1) explicitly locate related objects guided by textual entities to exclude undesired visual content, and 2) incrementally model the multi-turn structure of the dialogue to develop a unified representation combining multi-turn utterances with the corresponding related objects. However, previous work overlooks these two important aspects.


In this paper, we thus propose a novel and effective {\bf M}ultimodal {\bf I}ncremental {\bf T}ransformer with {\bf V}isual {\bf G}rounding, named MITVG, which contains two key parts: visual grounding and multimodal incremental transformer. Visual grounding aims to establish specific mapping of objects and textual entities by explicitly locating related objects in the image with the textual entities. By doing so, our model can exclude undesired visual content and reduce attention noise. On the basis of visual grounding, the multimodal incremental transformer is used to model the multi-turn dialogue history combined with the specific visual content to generate visually and contextually coherent responses. As an encoder-decoder framework, MITVG contains a Multimodal Incremental Transformer Encoder (MITE) and a Gated Cross-Attention Decoder (GCAD).

We test the effectiveness of our proposed model on large-scale datasets: VisDial v0.9 and v1.0~\cite{das2017visual}. Both automatic and manual evaluations show that our model substantially outperforms the competitive baselines and achieves the new state-of-the-art results on substantial metrics.
Our main contributions are as follows:
\begin{itemize}
  \item To the best of our knowledge, we are the first to leverage visual grounding to explicitly locate related objects in the image guided by textual entities for visual dialogue.
  \item We propose a novel multimodal incremental transformer to encode the multi-turn dialogue history step by step combined with the visual content and then generate a contextually and visually coherent response.
  \item We achieve comparable performance on VisDial v0.9 and v1.0 datasets.
\end{itemize}


\section{Approach}

\begin{figure*}[t!]
\centering
\scalebox{0.87}{
  \begin{overpic}[width=\textwidth]{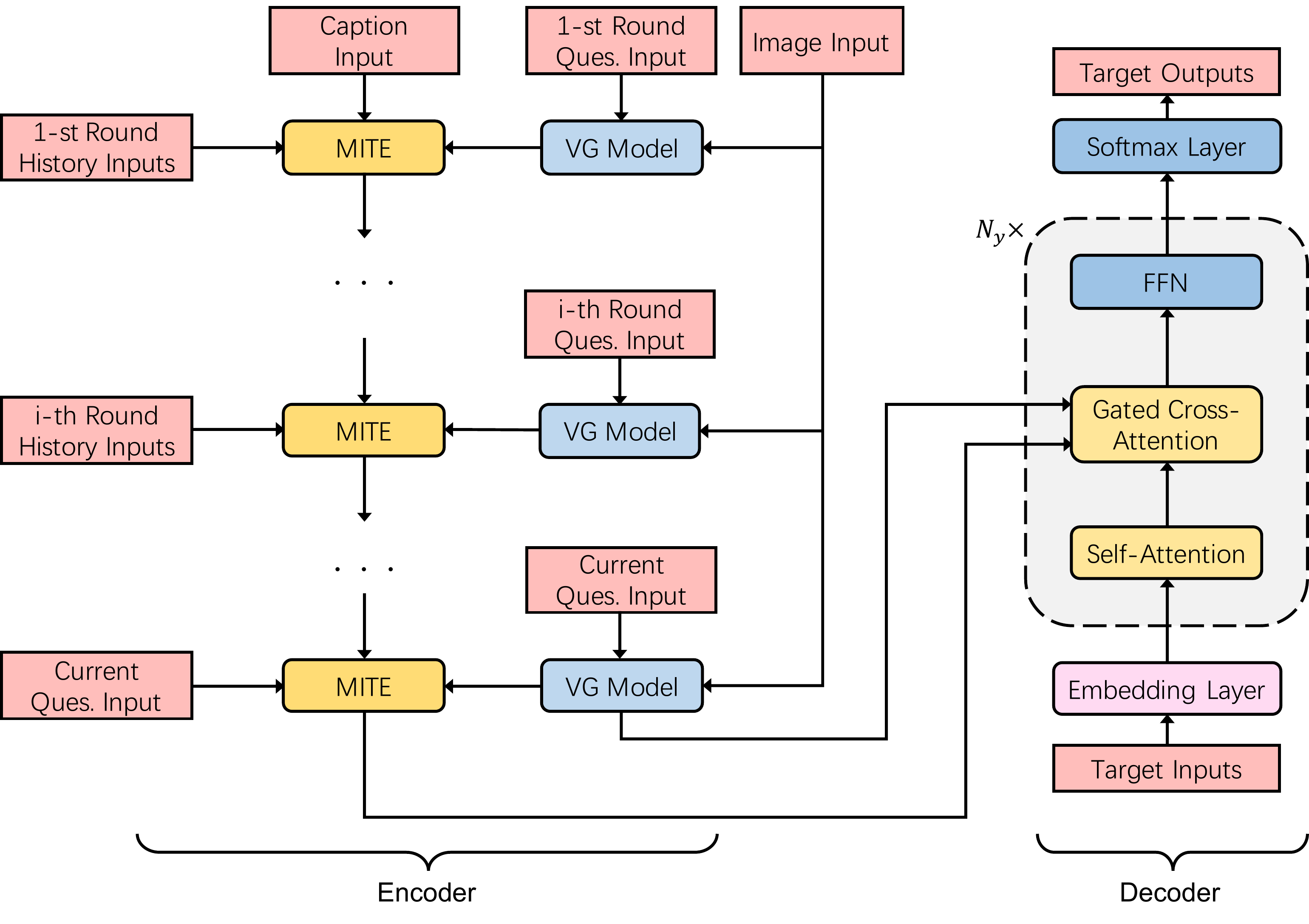}
  \end{overpic}
  }\vspace{-5pt}
  \caption{The framework of {\bf M}ultimodal {\bf I}ncremental {\bf T}ransformer with {\bf V}isual {\bf G}rounding (MITVG). ``VG Model'' indicates visual grounding model~\cite{yang2019fast} (Details are described in Sec.~\ref{sec:vg}). ``MITE'' denotes the multimodal incremental transformer encoder (Details are described in Sec.~\ref{sec:MITE}). MITVG firstly uses the VG model to explicitly model the relationship between the textual content and the visual content, and encodes multi-turn dialogue history in the order of the dialogue based on visual grounding, and finally utilizes the outputs of both encoder and visual grounding to generate the response word by word in the decoding process. 
  }\label{fig:model} 
\end{figure*}

\subsection{Overview}
In this section, we formally describe the visual dialogue task and then proceed to our proposed {\bf M}ultimodal {\bf I}ncremental {\bf T}ransformer with {\bf V}isual {\bf G}rounding (MITVG). 

Following \citeauthor{das2017visual}\shortcite{das2017visual}, a visual dialogue agent is given three inputs, i.e., an image $I$, a dialogue history (the caption and question-answer pairs) till round $t-1$: $H=(\underbrace{Cap}_{H_0}, \underbrace{(Q_1, A_1)}_{H_1}, \cdots, \underbrace{(Q_{t-1},A_{t-1})}_{H_{t-1}})$ and the current question $Q_t$ at round $t$, where $Cap$ is the caption describing the image taken as $H_0$ and $H_1, \dots, H_{t-1}$ are concatenations of question-answer pairs. The goal of the visual dialogue agent is to generate a response (or answer) $A_t$ to the question $Q_t$. $Cap$, $Q_*$ and $A_*$ are sentences.

\figref{fig:model} shows the framework of MITVG, which aims to explicitly model multi-turn dialogue history step by step based on the explicit modeling relationship between multiple modalities. MITVG firstly locates related objects in the image explicitly guided by the textual entities via visual grounding, then encodes multi-turn dialogue history in the order of the dialogue utterance based on visual grounding via Multimodal Incremental Encoder (MITE), and finally utilizes the outputs of both encoder and visual grounding to generate the response word by word via Gated Cross-Attention Decoder (GCAD). 

\subsection{Input Representation}
Before describing our method, we introduce the input representation.
\paragraph{Image Features.} 
We use a pre-trained Faster R-CNN model~\cite{ren2015faster} to extract object-level image features. Specifically, the image features $v$ for an image $I$ are represented by:
\begin{equation}
  v = {\rm Faster\ R-CNN}(I) \in \mathbb{R}^{K \times V} \label{v1},
\end{equation} 
where $K$ denotes the total number of the detected objects per image and $V$ denotes the dimension of features for each object.

\paragraph{Language Features.}
The current (at the $t$-th round) $L$-word question features are a sequence of $M$-dimension word embedding with positional encoding added ~\cite{vaswani2017attention}, as follows:
\begin{eqnarray}
  q_t &=& [s_{t,1}, s_{t,2}, \dots, s_{t,L}] \in \mathbb{R}^{L \times M}, \label{eq:u1}\\
  s_{t,j} &=& w_j + PE(j), 
\end{eqnarray}
where $w_j$ is the word embedding of the $j$-th word in the question $Q_t$, and $PE(\cdot)$ denotes positional encoding function~\cite{vaswani2017attention}. For the dialogue history $H = \{H_0, H_1, \dots, H_{t-1}\}$ and the answer $A_t$, the dialogue history features $u = \{u_0, u_1, \dots, u_{t-1}\}$ and the answer features $a_t$ are obtained in the same way as the question $Q_t$.

\subsection{Visual Grounding}\label{sec:vg}
\begin{figure}[t!]
\centering
\scalebox{0.95}{
  \begin{overpic}[width=\columnwidth]{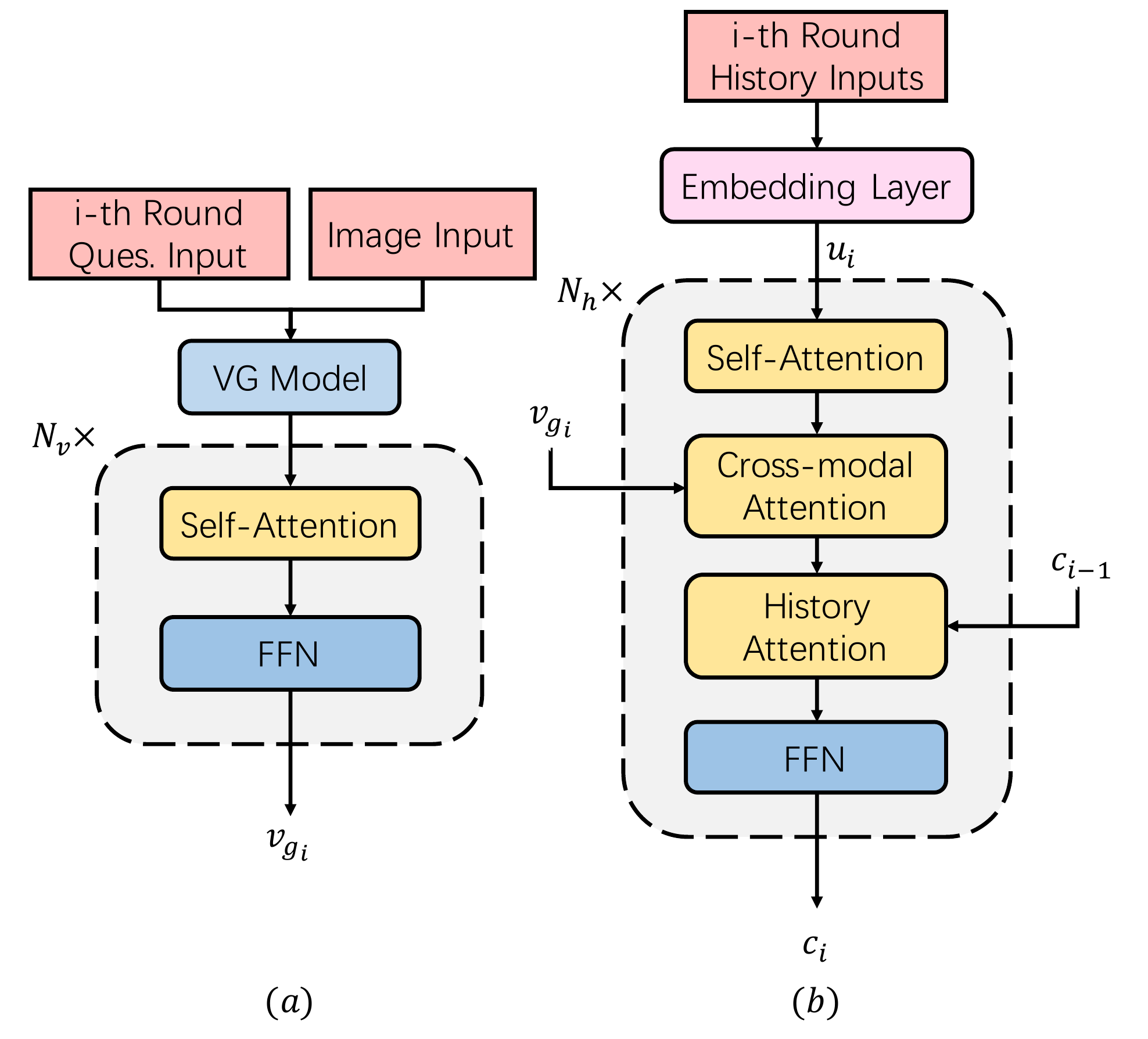}
  \end{overpic}
  } \vspace{-10pt}
  \caption{Framework of (a) Visual Grounding and (b) Multimodal Incremental Transformer Encoder (MITE).
  }\label{fig:MITE}
\end{figure}

To exclude the needless visual content, we introduce visual grounding, which is defined to ground a natural language query (phrase or sentence) about an image onto a correct region of the image. First of all, we use NeuralCoref\footnote{Introduction and code of NeuralCoref are available at https://github.com/huggingface/neuralcoref. NeuralCoref is only used for visual grounding.} for reference resolution. For example, when it processes the question Q4 ``{\em are they 
wearing matching uniforms ?}'' shown in \figref{fig:example}, NeuralCoref takes the question Q4 and its history as inputs, and then generates a new question ``{\em are the people wearing matching uniforms ?}'' as a new Q4.  As shown in \figref{fig:MITE} $(a)$, visual grounding model~\cite{yang2019fast} takes the $i$-th question $Q_i$ and the image $I$ as inputs and generates initial visual grounding features, as follows:
\begin{equation}
  {v}_{g_i}^{(0)} = {\rm VGM}(Q_i, I), \label{eq:vgm}\\
\end{equation}
where ${\rm VGM(\cdot)}$ denotes visual grounding model\footnote{Introduction and code are available at https://github.com/zyang-ur/onestage\_grounding.}. Then ${v}_{g_i}^{(0)}$ is sent to the multi-head self-attention layer followed by a position wise feed-forward network (FFN) layer (stacked $N_v$ times) to generate the $i$-th visual grounding features as follows\footnote{For simplicity, we omit the descriptions of layer normalization and residual connection.}: 
\begin{equation}
  \hat{v}_{g_i}^{n} = {\rm MultiHead}\left(v_{g_i}^{(n-1)}, v_{g_i}^{(n-1)}, v_{g_i}^{(n-1)}\right), \label{eq:u1}
\end{equation}
where $n = 1, \dots, N_v$ and ${\rm MultiHead}(\cdot)$ denotes the multi-head self-attention layer~\cite{vaswani2017attention}, then
\begin{equation}
 v_{g_i}^{(n)} = {\rm FFN}\left(\hat{v}_{g_i}^{n}\right), 
 \end{equation}
where $n = 1, \dots, N_v$ and ${\rm FFN}(\cdot)$ denotes the position wise feed-forward networks~\cite{vaswani2017attention}. After $N_v$ layers computation, we obtain the final visual grounding features $v_{g_i}$ by:
\begin{eqnarray}
 v_{g_i} &=& v_{g_i}^{(N_v)},
\end{eqnarray}
Actually, there are some questions that do not contain any entities in the visual dialogue, such as ``{\em anything else ?}''. For such questions, we use the features of the whole image instead, i.e. $v_{g_i} = v$.

\subsection{Multimodal Incremental Transformer}
Inspired by the idea of incremental transformer~\cite{li2019incremental} which is originally designed for the single-modal dialogue task, we make an extension and propose a multimodal incremental transformer, which is composed of a Multimodal Incremental Transformer Encoder (MITE) and a Gated Cross-Attention Decoder (GCAD). The MITE uses an incremental encoding scheme to encode multi-turn dialogue history with an understanding of the image. The GCAD leverages the outputs from both the encoder and visual grounding via the gated cross-attention layer to fuse the two modal information in order to generate a contextually and visually coherent response word by word.

\subsubsection{MITE}\label{sec:MITE}
To effectively encode multi-turn utterances grounded in visual content, we design the Multimodal Incremental Transformer Encoder (MITE). As shown in \figref{fig:MITE} $(b)$, at the $i$-th round, where $i=1,2,...,t-1$, the MITE takes the visual grounding features $v_{g_i}$, the dialogue history features $u_i$ and the context state $c_{i-1}$ as inputs, and utilizes attention mechanism to incrementally build up the representation of the relevant dialogue history and the associated image regions, and then outputs the new context state $c_i$. This process can be stated recursively as follows:
\begin{equation}
  c_i = {\rm MITE}\left(v_{g_i}, u_i, c_{i-1}\right),
\end{equation} 
where ${\rm MITE(\cdot)}$ denotes the encoding function, $c_i$ denotes the context state after the dialogue history features $u_i$ and the visual grounding features $v_{g_i}$ being encoded, and $c_0$ is the dialogue history features $u_0$.

As shown in \figref{fig:MITE} $(b)$, we use a stack of $N_h$ identical layers to encode $v_{g_i}$, $u_i$ and $c_{i-1}$, and to generate $c_i$. Each layer consists of four sub-layers. 
{\bf The first sub-layer} is a multi-head self-attention for the dialogue history:
\begin{equation}
  {\rm A^{(n)}} = {\rm MultiHead}\left({\rm C}^{(n-1)}, {\rm C}^{(n-1)}, {\rm C}^{(n-1)}\right),
\end{equation} 
where $n = 1, \dots, N_h$, ${\rm C}^{(n-1)}$ is the output of the
last layer $N_{n-1}$, and ${\rm C}^{(0)}$ is the dialog history features $u_i$. 
{\bf The second sub-layer} is a multi-head cross-modal attention:
\begin{equation}
  {\rm B^{(n)}} = {\rm MultiHead}\left({\rm A}^{n}, v_{g_i}, v_{g_i}\right),
\end{equation} 
where $v_{g_i}$ is the visual grounding features. 
{\bf The third sub-layer} is a multi-head history attention:
\begin{equation}
  {\rm F^{(n)}} = {\rm MultiHead}\left({\rm B}^{(n)}, {\rm c}_{i-1}, {\rm c}_{i-1}\right),
\end{equation}
where $c_{i-1}$ is the context state after the previous dialogue history features $u_{i-1}$ being encoded. That’s why we call this encoder ``Multimodal Incremental Transformer''. 
{\bf The fourth sub-layer} is a position wise feed-forward network (FFN): 
\begin{equation}
  {\rm C^{(n)}} = {\rm FFN}\left({\rm F}^{(n)}\right).
\end{equation}
We use $c_i$ to denote the final representation at $N_h$-th layer:
\begin{equation}
  c_i = {\rm C}^{(N_h)} .
\end{equation}
The mulitmodal incremental transformer encoder at the current turn $t$, i.e., the bottom one in \figref{fig:model}, has the same structure as all the other MITEs but takes the visual grounding features $v_{g_t}$, the current question features $q_t$ and the context state $c_{t-1}$ as inputs and generates the final context state $c_t$.

\subsubsection{GCAD} \label{sec:gcad}
Motivated by the real-world human cognitive process, we design a Gated Cross-Attention Decoder (GCAD) shown in \figref{fig:model}, which takes the masked answer features $a_{<z}$ (where $z = 1,2,...,Z$ and $Z$ is the length of the answer), encoder outputs $c_t$ and visual grounding features $v_{g_t}$ as inputs, and generates contextually and visually coherent responses grounded in an image. GCAD is composed of a stack of $N_y$ identical layers, each of which has three sub-layers. 

{\bf The first sub-layer} is a multi-head self-attention as follows:
\begin{equation}
 {\rm J}^{(n)} = {\rm MultiHead}\left({\rm R}^{(n-1)}, {\rm R}^{(n-1)}, {\rm R}^{(n-1)}\right), \label{eq:J}
\end{equation}
where $n = 1, \dots, N_y$, ${\rm R}^{(n-1)}$ is the output of the previous layer, and ${\rm R}^{(0)}$ is the masked answer features $a_{<z}$. 

\begin{figure}[t!]
\centering
\scalebox{0.9}{
  \begin{overpic}[width=\columnwidth]{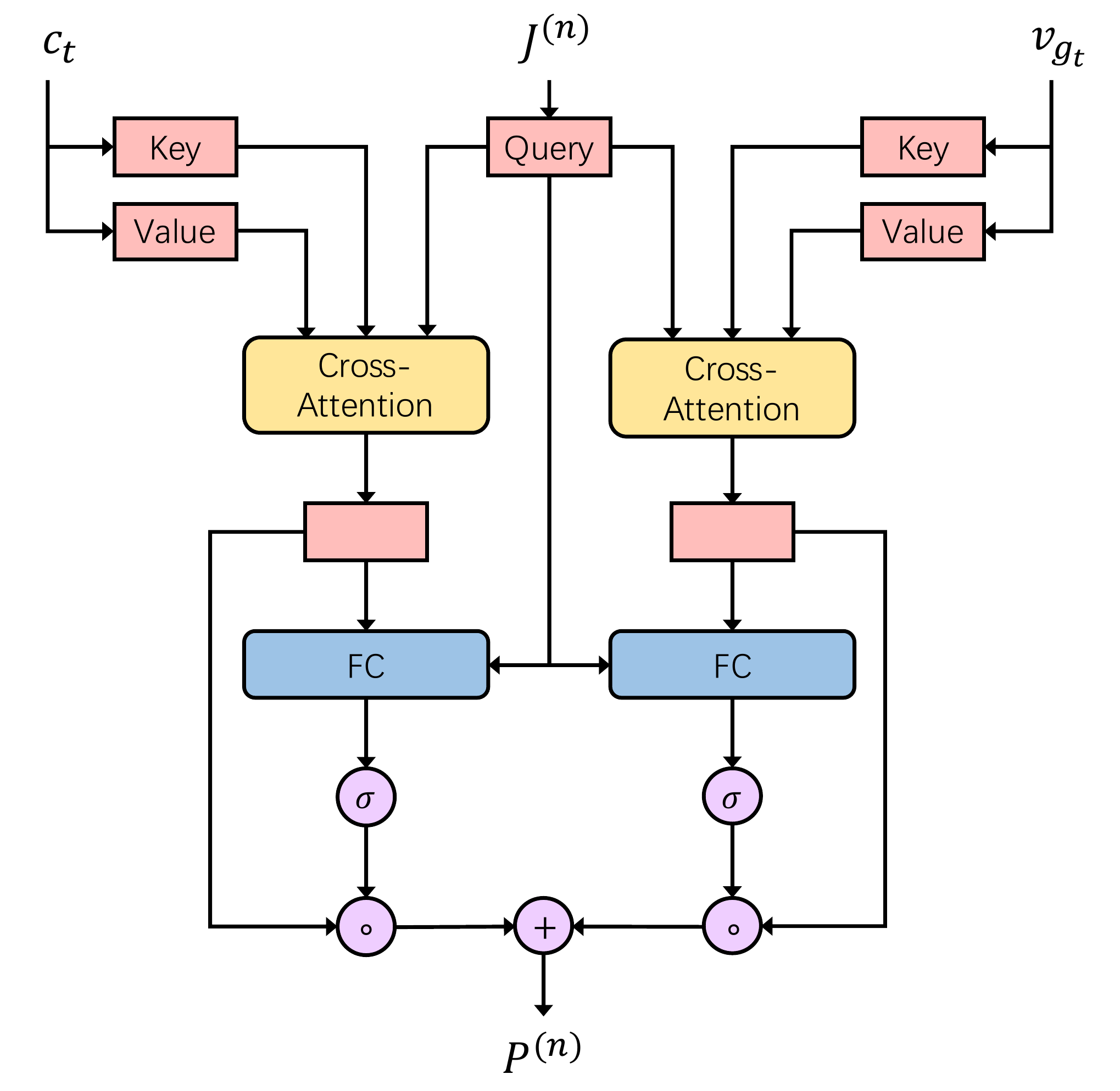}
  \end{overpic}
  }\vspace{-11pt}
  \caption{Framework of Gated Cross-Attention (GCA) in the Deocer. 
  }\label{fig:decoder}
  \vspace{-1em}
\end{figure}

{\bf The second sub-layer} is a multi-head gated cross-modal attention layer (GCA) as shown in \figref{fig:decoder}, calculated as:
\begin{equation}
 {\rm P}^{(n)} = \alpha^{(n)} \circ {\rm E}^{(n)} + \beta^{(n)} \circ {\rm G}^{(n)}, \label{eq:gca1}
\end{equation}
where $n = 1, \dots, N_y$, $\circ$ denotes Hadamard product, ${\rm E}^{(n)}$ and ${\rm G}^{(n)}$ denote the outputs of two cross-attention functions, computed as follows:
\begin{eqnarray}
  {\rm E}^{(n)} &=& {\rm MultiHead}\left({\rm J}^{(n)}, c_t, c_t\right),\\
  {\rm G}^{(n)} &=& {\rm MultiHead}\left({\rm J}^{(n)}, v_{g_t}, v_{g_t}\right),
\end{eqnarray}
where $\alpha^{(n)}$, $\beta^{(n)}$ are two gates\footnote{Our inspiration comes from~\citeauthor{cornia2020m2}~\shortcite{cornia2020m2}. }:
\begin{eqnarray}
  \alpha^{(n)} &=& \sigma\left(W_E[{\rm J}^{(n)}, E^{(n)}]+ b_E\right),\\
  \beta^{(n)} &=& \sigma\left(W_G[{\rm J}^{(n)}, G^{(n)}]+ b_G\right), \label{eq:gca2}
\end{eqnarray}
where  $\sigma$ denotes sigmoid function, $W_E$, $W_G$, $b_E$, $b_G$ are learnable parameters, and $[\cdot, \cdot]$ indicates concatenation. 

{\bf The third sub-layer} is a position wise feed-forward network (FFN): 
\begin{equation}
  {\rm R^{(n)}} = {\rm FFN}\left({\rm P}^{(n)}\right).
\end{equation}
We use $r_z$ to denote the final representation at $N_y$-th layer:
\begin{equation}
  r_z = {\rm R}^{(N_y)} .
\end{equation}
Finally, we use softmax to get the word probabilities ${\hat a}_z$:
\begin{equation}
  {\hat a}_z = {\rm softmax}(r_z).
\end{equation}

\section{Experiments}
\label{sec:Experiments}
\subsection{Datasets}
We conduct experiments on the VisDial v0.9 and v1.0 datasets~\cite{das2017visual} to verify our approach.
VisDial v0.9 contains 83k dialogs on COCO-train~\cite{lu2017best} and 40k dialogs on COCO-val images as test set, for a total of 1.23M dialog question-answer pairs. VisDial v1.0 dateset is an extension of VisDial v0.9 dateset with additional 10k COCO-like images from Flickr. VisDial v1.0 dateset contains 123k, 2k and 8k images as train, validation and test splits, respectively.

\begin{table*}[h]
  \centering
    \resizebox{0.74\textwidth}!{
  \begin{tabular}{l|c|c|c|c|c|c|c}
    \toprule
    Model & Object & Vis-G & MRR $\uparrow$ & R@1 $\uparrow$ & R@5$\uparrow$ & R@10 $\uparrow$ & Mean $\downarrow$ \\
    \midrule
    AP~\cite{das2017visual}& $\times$ & $\times$& 37.35 & 23.55 & 48.52 & 53.23 & 26.50 \\
    NN~\cite{das2017visual} & $\times$ & $\times$& 42.74 & 33.13 & 50.83 & 58.69 & 19.62 \\
    LF~\cite{das2017visual} & $\times$ & $\times$ & 51.99 & 41.83 & 61.78 & 67.59 & 17.07\\
    HREA~\cite{das2017visual}& $\times$  & $\times$ & 52.42 & 42.28 & 62.33 & 68.71 & 16.79 \\
    MN~\cite{das2017visual}& $\times$ & $\times$ & 52.59 & 42.29 & 62.85 & 68.88 & 17.06 \\
    HCIAE~\cite{lu2017best}& $\times$ & $\times$ & 53.86 & 44.06 & 63.55 & 69.24 & 16.01 \\
    CorefNMN~\cite{Kottur2018VisualCR} & $\times$ & $\times$ & 53.50 & 43.66 & 63.54 & 69.93 & 15.69 \\
    CoAtt~\cite{wu2018you} & $\times$ & $\times$ & 55.78 & 46.10 & 65.69 & 71.74 & 14.43 \\
    RvA~\cite{niu2019recursive} & $\checkmark$ & $\times$ & 55.43 & 45.37 & 65.27 & \underline{72.97} & \underline{\bf{10.71}} \\
    DVAN~\cite{guo2019dual} &  $\checkmark$ &  $\times$ & 55.94 & 46.58 & 65.50 & 71.25 & 14.79 \\
    VDBERT~\cite{wang2020vd}  &  $\checkmark$ & $\times$ & 55.95 & \underline{46.83} & 65.43 & 72.05 & 13.18 \\
    LTMI~\cite{nguyenefficient}$^\dagger$ &  $\checkmark$ & $\times$ & 55.85 & 46.07 & 65.97 & 72.44 & 14.17 \\
    DMRM~\cite{chen2020dmrm}  & $\checkmark$ & $\times$ & \underline{55.96} & 46.20 & \underline{66.02} & 72.43 & 13.15\\
    \midrule
    MITVG  & $\checkmark$  & $\checkmark$ & \bf{56.83} & \bf{47.14} & \bf{67.19} & \bf{73.72} & 11.95\\
    \bottomrule
  \end{tabular}
}
  \caption{Performance on VisDial val v0.9~\cite{das2017visual}. 
  $\dagger$ indicates that we re-implement the model. ``Object'' and ``Vis-G'' denote if the model uses object-level image features and visual grounding, respectively. Underline denotes the highest score among baselines. Our MITVG exceeds previous work on most of the metrics and achieves comparable performance. 
  }\label{tab:resultv0.9} \vspace{-5pt}
  \end{table*}
  
  \begin{table*}[h]
  \centering
    \resizebox{0.80\textwidth}!{
  \begin{tabular}{l|c|c|c|c|c|c|c|c}
    \toprule
    Model  & Object & Vis-G & MRR $\uparrow$ & R@1 $\uparrow$ & R@5$\uparrow$ & R@10 $\uparrow$& Mean $\downarrow$ & NDCG $\uparrow$\\
    \midrule
    MN~\cite{das2017visual}$^\ddagger$  & $\checkmark$ & $\times$ & 47.99 & 38.18 & 57.54 & 64.32 & 18.60 & 51.86\\
    HCIAE~\cite{lu2017best}$^\ddagger$  & $\checkmark$ & $\times$ & 49.07 & 39.72 & 58.23 & 64.73 & 18.43 & 59.70\\
    CoAtt~\cite{wu2018you}$^\ddagger$ &$\checkmark$  & $\times$ & 49.64 & 40.09 & 59.37 & 65.92 & 17.86 & 59.24 \\
    Primary~\cite{guo2019image}  & $\checkmark$ & $\times$ & 49.01 & 38.54 & 59.82 & 66.94 & 16.60 & -\\
    ReDAN~\cite{gan2019multi} & $\checkmark$ & $\times$ & 50.02 &  40.27 & 59.93 & 66.78 & 17.40 &  60.47 \\
    DMRM~\cite{chen2020dmrm} &$\checkmark$ & $\times$ &  50.16 & 40.15 & 60.02 & 67.21 & \underline{15.19} & -  \\
    LTMI~\cite{nguyenefficient}$^\dagger$ &$\checkmark$ & $\times$ & 50.38 & 40.30 & 60.72 & \underline{68.44} & 15.73  & \underline{61.61}  \\
    DAM~\cite{jiang2020dam} &$\checkmark$ & $\times$ &  \underline{50.51} & \underline{40.53} & \underline{60.84} & 67.94 & 16.65 & 60.93  \\
    KBGN~\cite{jiang2020kbgn}&$\checkmark$ & $\times$ & 50.05 & 40.40 & 60.11 & 66.82 & 17.54 & 60.42\\
    \midrule
    MITVG & $\checkmark$& $\checkmark$ & \bf{51.14}   & \bf{41.03} & \bf{61.25} & \bf{68.49} & \bf{
14.37} & 61.47 \\
    \bottomrule
  \end{tabular}
}
  \caption{Performance on VisDial val v1.0~\cite{das2017visual}. $\ddagger$ denotes that all the models are re-implemented by \citeauthor{gan2019multi}~\shortcite{gan2019multi}. Our MITVG outperforms previous work and achieves comparable performance. 
  }\label{tab:resultv1.0} \vspace{-8pt}
  \end{table*}

\subsection{Implementation and Evaluation}
\paragraph{Implementation Details.}
Following previous work~\cite{das2017visual}, in order to represent words we firstly lowercase all the texts and convert digits to words, and then remove contractions before tokenization. The captions, questions and answers are further truncated to ensure that they are not longer than 40, 20 and 20 tokens, respectively. We construct the vocabulary of tokens that appear at least 5 times in the training split. To represent image regions, we use Faster R-CNN~\cite{ren2015faster} with ResNet-101~\cite{he2016deep} finetuned on the Visual Genome dataset~\cite{krishna2017visual}, thus obtaining a 2048-dimensional feature vector for each region. The layers of our encoder, decoder and visual grounding module are all set to 3. The number of attention heads in multi-head attention is 8 and the filter size is 2048. The word embedding is shared by the history, questions and responses. The dimension of word embedding is set to 512 empirically. We use Adam~\cite{kingma2014adam} for optimization, following the learning rate scheduling strategy of \citeauthor{vaswani2017attention}~\shortcite{vaswani2017attention}. Our model is implemented using PyTorch v1.0, Python v3.6, and provides out of the box support with CUDA 9 and CuDNN 7.
We train our model on TITAN XP with 8 GPUs. For each epoch, we spend about 9,000 seconds on training the model. The total parameters are about 56.79M.

Before we train our model, we use three external tools for image features extracting, reference resolution and visual grounding.

\paragraph{Image Features Extracting} We extract image features of VisDial images, using a Faster-RCNN~\cite{ren2015faster} with ResNet-101~\cite{he2016deep} pre-trained on Visual Genome~\cite{krishna2017visual}, introduction and code from https://github.com/peteanderson80/bottom-up-attention.

\paragraph{Reference Resolution} we use NeuralCoref v4.0 for reference resolution, which is developed by huggingface. Introduction and code are available at https://github.com/huggingface/neuralcoref.

\paragraph{Visual Grounding} We use One-Stage Visual Grounding Model~\cite{yang2019fast} to obtain the visual grounding features. Introduction and code are available at https://github.com/zyang-ur/onestage\_grounding.


\paragraph{Automatic Evaluation.} We use a retrieval setting to evaluate individual responses at each round of a dialogue, following~\citet{das2017visual}. Specifically, at test time, apart from the image, ground truth dialogue history and the question, a list of 100-candidate answers is also given. The model is evaluated on retrieval metrics: (1) rank of human response (Mean, the lower the better), (2) existence of the human response in $top-k$ ranked responses, i.e., R@$k$ (3) mean reciprocal rank (MRR) of the human response (the higher the better) and (4) normalized discounted cumulative gain (NDCG) for VisDial v1.0 (the higher the better). During evaluation, we use the log-likelihood scores to rank candidate answers.

\paragraph{Human Evaluation.} We randomly extract 100 samples for human evaluation according to \citet{wu2018you}, and then ask 3 human subjects to guess whether the last response in the dialogue is human-generated or machine-generated. If at least 2 of them agree it is generated by a human, we think it passes the Truing Test (M1). In addition, we record the percentage of responses that are evaluated better than or equal to human responses (M2), according to the human subjects’ evaluation.

\subsection{Main Results}
We compare our proposed model to the state-of-the-art {\em generative models} developed in previous work. Current encoder-decoder based generative models can be divided into tree facets. (1) Fusion-based models: LF~\cite{das2017visual} and HREA~\cite{das2017visual} directly encode the multimodal inputs and decode the answer. (2) Attention-based models: HCIAE~\cite{lu2017best}, CoAtt~\cite{wu2018you}, Primary~\cite{guo2019image}, ReDAN~\cite{gan2019multi}, DVAN~\cite{guo2019dual} and DMRM~\cite{chen2020dmrm}, DAM, LTMI, KBGN. (3) Visual co-reference resolution models: CorefNMN~\cite{Kottur2018VisualCR}, RvA~\cite{niu2019recursive}. (4) The pretraining model: VDBERT~\cite{wang2020vd}.


As shown in \tabref{tab:resultv0.9} and \tabref{tab:resultv1.0}, our MITVG, which explicitly locates related objects guided by the textual entities and implements a multimodal incremental transformer to incrementally build the representation of the dialogue history and the image, achieves comparable performance on the VisDial v0.9 and v1.0 datasets. Specifically, our model outperforms previous work by a significant margin both on the VisDial v0.9 dataset (0.87 on MRR, 0.31 on R@1, 1.17 on R@5, 0.75 on R10) and the VisDial
v1.0 dataset (0.98 on MRR, 0.76 on R@1, 1.23 on R@5, 1.28 on R10, 0.82 on Mean, and 1.00 on NDCG). 
The improvement of R@10 is the largest and our method also gains a large increase on MRR and R@1 due to the explicit modeling of multiple modalities (Seeing Sec~\ref{sec:casestudy} for further quantitative analysis).

As shown in~\tabref{tab:humanstudy}, we conduct human study to further prove the effectiveness of our model. Our model achieves the highest scores both on the metric M1 (0.76) and M2 (0.70) compared with the previous model, DMRM~\cite{chen2020dmrm}. These results show that our model can generate a better contextually and visually coherent response. 


\begin{table}
  \centering
    \resizebox{0.70\columnwidth}!{
  \begin{tabular}{p{2.5cm}|c|c}
    \toprule
      & DMRM & MITVG\\
    \midrule
    Method 1 (M1) &  0.62 & \bf{0.76}\\
    \midrule
    Method 2 (M2) & 0.59 & \bf{0.70}\\
    \bottomrule
  \end{tabular}
  }
  \caption{Human evaluation on 100 sampled responses on VisDial val v1.0.  M1: percentage of responses pass the Turing Test. M2: percentage of responses evaluated better than or equal to human responses. }\label{tab:humanstudy} \vspace{-10pt}
\end{table}

\subsection{Ablation Study}
We also conduct an ablation study to illustrate the validity of our proposed Multimodal Incremental Transformer with Visual Grounding. The results are shown in~\tabref{tab:ablationstudy}.

We implement Multimodal Incremental Transformer without Visual Grounding (`MITVG w/o VG') to verify the validity of visual grounding. As shown in \tabref{tab:ablationstudy}, comparing `MITVG w/o VG' with MITVG, we find the metrics decrease obviously (0.46 on MRR, 0.60 on R@1, 0.68 on R@5, 0.46 on R@10 and 0.59 on Mean) if visual grounding is deleted from MITVG. This observation demonstrates the validity of visual grounding.

To verify the effectiveness of the incremental transformer architecture, we implement a Multimodal Incremental LSTM without Visual Grounding (`MI-LSTM w/o VG'). A 3-layer bidirectional LSTM~\cite{schuster1997bidirectional} with multi-head attention and a 1-layer LSTM with GCA are applied for encoder and decoder, respectively. All the LSTM hidden state size is 512.
Results in \tabref{tab:ablationstudy} demonstrate the effectiveness of our incremental transformer architecture (compare `MITVG w/o VG' with `MI-LSTM w/o VG'). Results from the comparison between `MITVG w/o VG' and DMRM~\cite{chen2020dmrm} also show the validity of our incremental transformer to some extent.



\begin{table}
  \centering
  \resizebox{1.0 \columnwidth}!{
  \begin{tabular}{l|c|c|c|c|c}
    \toprule
    Model & MRR  & R@1  & R@5  & R@10 & Mean \\
    \midrule
    DMRM & 50.16 & 40.15 & 60.02 & 67.21 & 15.19 \\
    \midrule
    MITVG & \bf{51.14} & \bf{41.03} & \bf{61.25} & \bf{68.49} & \bf{
14.37} \\
    MITVG w/o VG & 50.68 & 40.43 & 60.57 & 68.03 & 14.96\\
    MI-LSTM w/o VG & 50.02 & 39.85 & 59.86 & 67.16 & 15.78\\
    \bottomrule
  \end{tabular}
  }
  \caption{Ablation study of our proposed model on VisDial val v1.0. ``MI-LISM'' indicates Multimodal Incremental LSTM. ``VG'' indicates visual grounding.}\label{tab:ablationstudy} \vspace{-5pt}
  \end{table}
  
\begin{table}
  \centering
    \resizebox{0.70\columnwidth}!{
  \begin{tabular}{c|c|c|c}
    \toprule
      & Train & Validation & Test\\
    \midrule
    VisDial v0.9 & 2.04 & 1.95 & -\\
    \midrule
    VisDial v1.0 & 2.05 & 1.93 & 1.93\\
    \bottomrule
  \end{tabular}
  }
  \caption{Average number of the grounded objects in each question.}\label{tab:Boxes} \vspace{-5pt}
\end{table}
  
\begin{figure*}[t!]
    \centering
    \scalebox{0.95}{
  \begin{overpic}[width=0.98\textwidth]{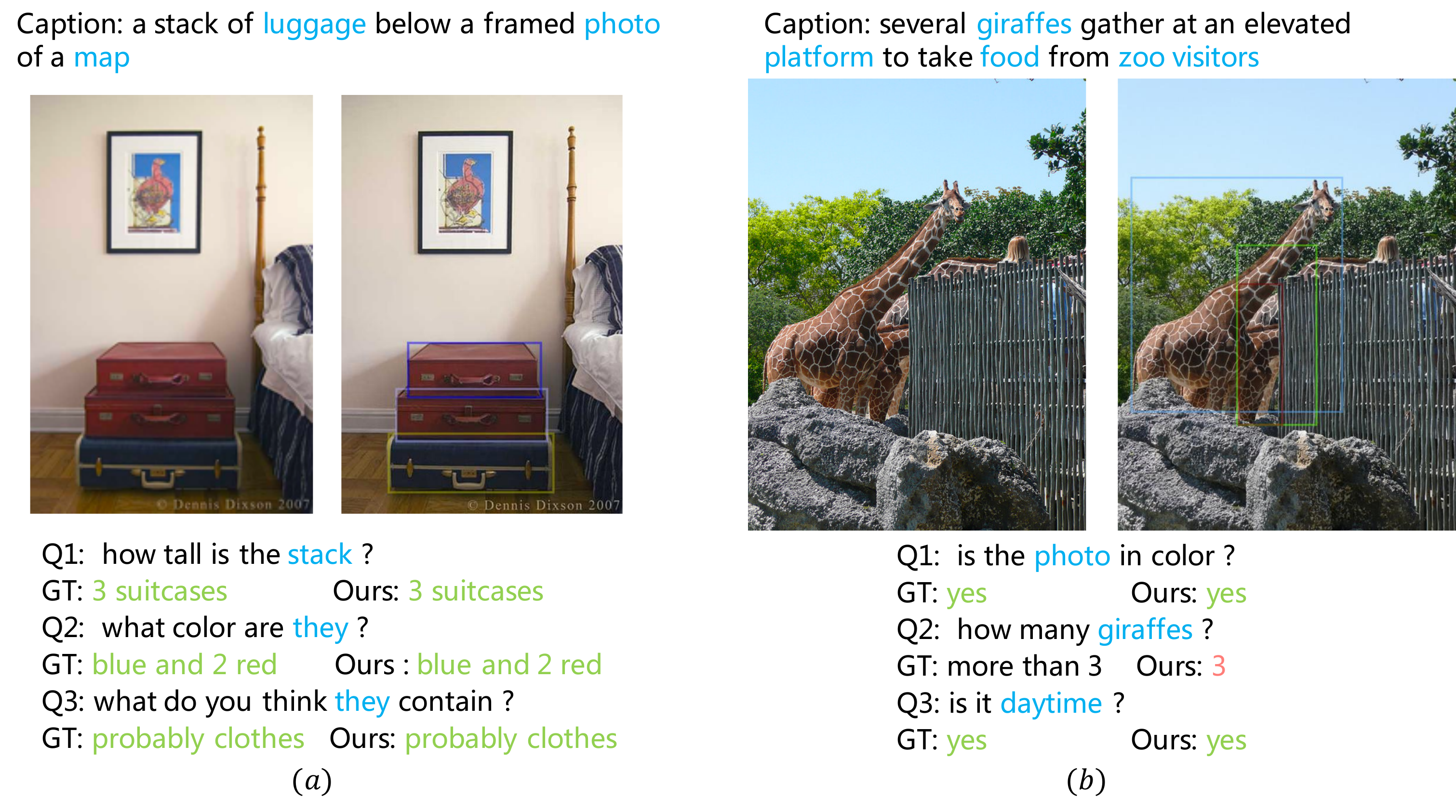} 
  \end{overpic}
  } 
  \caption{Case study. The text marked in blue indicates the dialogue topic. The answers marked in green and red indicate the right and wrong answers, respectively. Our MITVG often generates right responses (marked in green) in keeping with human answers.
  }\label{fig:case} 
\end{figure*}
  
\subsection{Case Study} \label{sec:casestudy}
As shown in \tabref{tab:Boxes}, we calculate the average number of the objects associated with entities in each question for assistant analysis.
As shown in \figref{fig:case} $(a)$, owing to the explicit understanding of visual content via visual grounding and the multimodal incremental transformer architecture, our MITVG generates responses in keeping with human answers. For example, while answering the question Q1 `{\em `how tall is the stack ?}'' and Q2 ``{\em what color are they ?}'', our model grounds the three suitcases accurately via visual grounding, thus giving the accurate responses ``{\em 3 suitcases}'' and ``{\em blue and 2 red}''. However, as shown in \figref{fig:case} $(b)$, for questions Q2, MITVG gives a wrong answer because it focuses on wrong number of objects in the question by visual grounding.




\section{Related Work} 
\paragraph{Visual Dialogue.}
Our work touches two branches of the research in visual dialogue. One is how to leverage image features. \citet{niu2019recursive} utilize object-level image features as visual attention and refine it by recursively reviewing the dialog history.  \citet{gan2019multi} and \citet{chen2020dmrm} regard the object-level image features as visual memory to infer answers progressively through multiple steps. The other is how to model dialogue history. \citet{Yang2019ICCV} propose a new training paradigm inspired by actor-critic policy gradient~\cite{sutton1999policy} for history-advantage training. \citet{guo2020iterative} represent each turn dialogue history with visual content as a node in a context-aware graph neural network. \citet{park2020multi} refine history information from both topic aggregation and context matching. Different from these approaches, we explicitly establish specific mapping of objects and textual entities to exclude undesired visual content via visual grounding, and model multi-turn structure of the dialogue based on visual grounding to develop a unified representation combining multi-turn utterances along with the relevant objects.

\paragraph{Incremental Structures.}
There are some successes on introducing the incremental structure into tasks related to dialog systems~\cite{zilka2015incremental,coman2019incremental,li2019incremental,das2017visual}. In particular, \citet{coman2019incremental} propose an incremental dialog state tracker which is updated on a token basis from incremental transcriptions. \citet{li2019incremental} devise an incremental transformer to encode multi-turn utterances along with knowledge in related documents for document grounded conversations. 
\citet{das2017visual} propose a dialog-RNN to produce an encoding for this round and a state for next round. Our model is different from these approaches mainly in two aspects: 1) we explicitly model the relationship between modalities, i.e., textual utterance and image objects, in visual dialogue through visual grounding; 2) based on the explicit association between modalities, our model incrementally encodes the dialogue history and the image with well-designed incremental multimodal architecture to sufficiently understand the dialogue content, thus generating better responses.

\section{Conclusion}
We propose a novel Multimodal Incremental Transformer with Visual Grounding for visual dialogue, named MITVG, which consists of two key parts: visual grounding and multimodal incremental transformer. Visual grounding aims to explicitly model the relationship between multiple modalities. Based on visual grounding, multimodal incremental transformer aims to explicitly model multi-turn dialogue history in the order of the dialogue. Experiments on the VisDial v0.9 and v1.0 datasets show that our model achieves comparable performance.
\bibliographystyle{acl_natbib}
\bibliography{anthology,acl2021}






\end{document}